\documentclass[10pt,twocolumn,letterpaper]{article}

\usepackage{wacv}
\usepackage{times}
\usepackage{epsfig}
\usepackage{graphicx}
\usepackage{amsmath}
\usepackage{amssymb}
\usepackage{booktabs}
\usepackage{comment}
\usepackage{bm}
\usepackage{caption}
\usepackage{subcaption}
\usepackage{multirow}
\usepackage[accsupp]{axessibility}

\DeclareMathOperator*{\argmax}{\arg\!\max}
\DeclareMathOperator*{\argmin}{\arg\!\min}
\newcommand{\vb}[1]{{\bm{#1}}}

\def\wacvPaperID{712} % Enter the WACV Paper ID here

\wacvalgorithmstrack   % Uncomment this line if you are submitting to the Algorithms Track.
\wacvfinalcopy % *** Uncomment this line for the final submission

\ifwacvfinal
\usepackage[breaklinks=true,bookmarks=false, colorlinks, citecolor=blue]{hyperref}
\else
\usepackage[pagebackref=true,breaklinks=true,colorlinks,bookmarks=false]{hyperref}
\fi

\begin{document}

\title{Scaling Novel Object Detection with \\Weakly Supervised Detection Transformers}

\author{Tyler LaBonte\textsuperscript{1,3}\quad
Yale Song\textsuperscript{2}\thanks{Work done at Microsoft Research, Redmond, WA.}\quad
Xin Wang\textsuperscript{1}\quad
Vibhav Vineet\textsuperscript{1}\quad
Neel Joshi\textsuperscript{1}\\
\textsuperscript{1}Microsoft Research \quad\textsuperscript{2}Meta AI, FAIR \quad\textsuperscript{3}Georgia Institute of Technology\\
\small\texttt{tlabonte@gatech.edu, yalesong@meta.com, \{wanxin, vivineet, neel\}@microsoft.com}
}

\maketitle
\thispagestyle{empty}

\begin{abstract}
A critical object detection task is finetuning an existing model to detect novel objects, but the standard workflow requires bounding box annotations which are time-consuming and expensive to collect. Weakly supervised object detection (WSOD) offers an appealing alternative, where object detectors can be trained using image-level labels. However, the practical application of current WSOD models is limited, as they only operate at small data scales and require multiple rounds of training and refinement. To address this, we propose the Weakly Supervised Detection Transformer, which enables efficient knowledge transfer from a large-scale pretraining dataset to WSOD finetuning on hundreds of novel objects. Additionally, we leverage pretrained knowledge to improve the multiple instance learning (MIL) framework often used in WSOD methods. Our experiments show that our approach outperforms previous state-of-the-art models on large-scale novel object detection datasets, and our scaling study reveals that class quantity is more important than image quantity for WSOD pretraining. The code is available at \href{https://github.com/tmlabonte/weakly-supervised-DETR}{this https URL.}
\end{abstract}

\section{Introduction}
\label{sec:introduction}

Object detection is a fundamental task in computer vision where supervised neural networks have demonstrated remarkable performance~\cite{RHG+15,RDG+16,LAE+16,CMS+20}. A major factor in the success of these approaches is the availability of datasets with fine-grained bounding box and segmentation annotations~\cite{EGW+10,LMB+14,KZG+17,GDG19,SLZ+19,KRA+20}. However, in comparison to image classification, the annotation process for object detection is considerably more expensive and time-consuming~\cite{PUK+17}. We consider weakly supervised object detection (WSOD), which aims to learn object detectors using only image-level category labels (\textit{i.e.}, classification labels).

Previous WSOD models~\cite{BV16,TWB+17} often generate object proposals using a low-precision high-recall heuristic~\cite{USG+13,ZD14}, then use multiple instance learning (MIL)~\cite{DLL97,ML98} to recover high-likelihood proposals. With proposal quality established as a major factor in object detection performance~\cite{HBD+15}, a practical direction is to leverage a \textit{source dataset} with bounding box annotations to transfer semantic (class-aware)~\cite{TWW+18b,CDZ+21} or class-agnostic~\cite{UPF18,ZWP+20} knowledge to a \textit{target dataset} of novel objects. These strategies enable the WSOD model to generate more accurate proposals and classifications by exploiting class and object similarity, respectively, between the source and target datasets.

\begin{figure*}[t]
\centering
\includegraphics[scale=0.35]{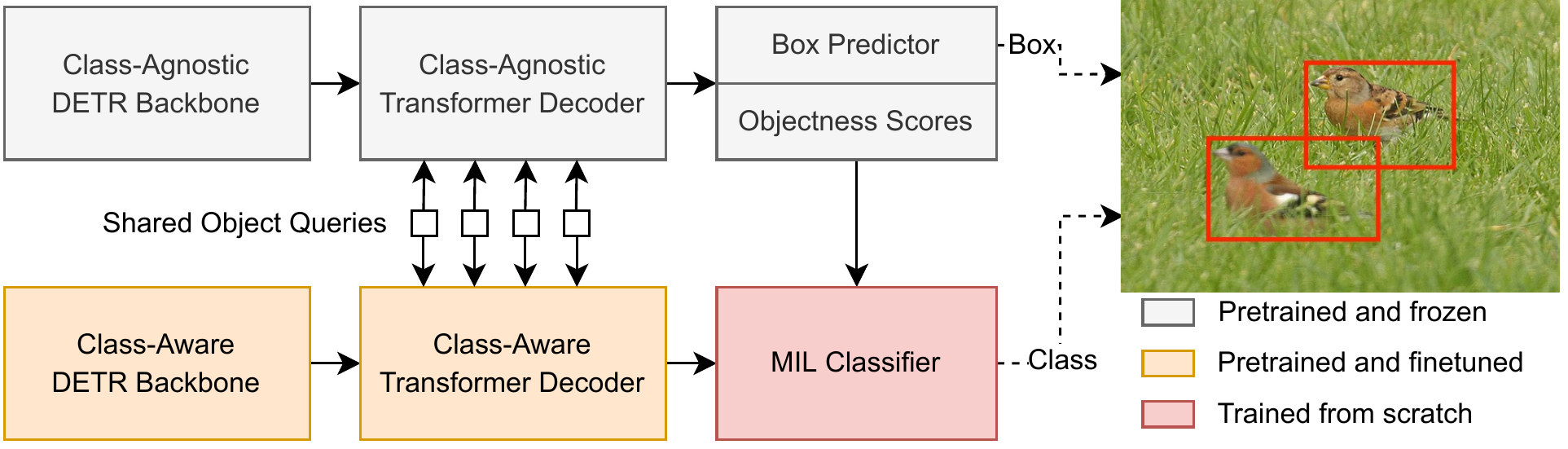}
\caption{WS-DETR is a hybrid approach utilizing a class-agnostic DETR as proposal generator and a class-aware DETR for weakly supervised finetuning. The two streams share object queries. The MIL classifier leverages objectness knowledge from the pretrained class-agnostic DETR to detect novel objects using only image-level labels.}
\label{fig:arch}
\end{figure*}

Though the presence of many classes in the source dataset is posited to be essential for effective transfer~\cite{UPF18}, current WSOD methods are typically designed for and trained on datasets with few classes. A widely-used setting in the literature is COCO-60~\cite{LMB+14,LKC18} (60 classes) to PASCAL VOC~\cite{EGW+10} (20 classes), with, to our knowledge, the largest source dataset being ILSVRC-179~\cite{RDS+15,ZWP+20} (179 classes) and the largest target dataset of novel objects being ILSVRC \texttt{val1b} (100 classes).\footnote{Uijlings \etal~\cite{UPF18} transfer from ILSVRC \texttt{val1a} (100 classes) to OpenImages~\cite{KRA+20} (600 classes), but with significant class overlap.} This has two major drawbacks which limit the usage of WSOD models in practice. First, knowledge transfer is most effective when objects in the target dataset have visually similar counterparts in the source dataset. In applications, training on few classes may limit the domains for which knowledge transfer is helpful. Second, current WSOD models perform best with multiple rounds of training and refinement~\cite{TWB+17,ZWP+20} or training an additional semantic model on the target dataset~\cite{CDZ+21}. In addition to the extra computation, these methods require a human to identify, \textit{e.g.}, the optimal number of refinements or pseudo ground truth mining steps, which is unscalable if there are hundreds of novel classes and many downstream tasks.

In contrast, what is desired in practice is similar to the pretraining-finetuning framework which has been standard in classification and fully supervised object detection~\cite{HZR+15,CLK+19} (though not in WSOD). Specifically, we would like to pretrain a single detection model on a large-scale, annotated source dataset with hundreds of classes, then use this model for weakly supervised finetuning on novel objects with image-level labels. Such a framework would empower practitioners to ``effortlessly'' solve WSOD tasks: the scale and diversity of the source dataset allows for knowledge transfer to a variety of distinct domains, while a simple, single-round finetuning procedure would enable computationally efficient WSOD training without the additional cost of semantic models or multi-round refinement.

To address this practical scenario, we propose the Weakly Supervised Detection Transformer (WS-DETR), which integrates DETR~\cite{CMS+20} with an MIL architecture for scalable WSOD finetuning on novel objects (detailed in Figure \ref{fig:arch}). A Transformer-based method is well-suited to this problem because, while they lack the inductive bias of CNNs, they excel at large-scale training and transfer learning for vision tasks~\cite{CMS+20,DBK+21}. Indeed, the Transformer~\cite{VSP+17} is the foundation for a widely-used machine learning workflow where massive pretrained models are finetuned to generalize to an incredible variety of downstream tasks, which aligns closely with our desired WSOD framework.

Existing MIL architectures \cite{BV16} are primarily based on a two-stage Faster R-CNN-like~\cite{RHG+15} structure where region-of-interest (RoI) pooling is performed on object proposals and the resultant features are passed to a classification head. Our WS-DETR is a hybrid which combines the comprehensive proposal generation of this two-stage framework with the scalability and simplicity of Transformer models.\footnote{Zhu \etal~\cite{ZSL+21} integrate DETR with a two-stage approach in the fully supervised setting; our methods and goals are not directly comparable.} We pretrain a class-agnostic DETR on the source dataset with binary object labels to serve as the proposal generator, then initialize a model with a frozen box predictor from a class-aware DETR pretrained on detection annotations (including classes) for WSOD training. Instead of RoI pooling, we utilize the DETR's learned positional embedding, also called object queries -- upon initialization, the object queries are set equal to the class-agnostic object queries and frozen, so the Transformer decoder attends to the same locations as the class-agnostic model and produces classification predictions which correspond to object proposals.

Additionally, we leverage pretrained knowledge to improve MIL training in two ways. First, we show that the objectness regularization of ~\cite{ZWP+20}, the state-of-the-art technique for incorporating the object proposal score of the pretrained class-agnostic model, falls prey to the well-known MIL weakness of overfitting to distinctive classification features of novel objects~\cite{TWB+17}. We propose a new formulation for estimating the joint object and class probabilities which rectifies this error without the need for a box refinement step~\cite{TWB+17}. Second, since the proposal generator outputs hundreds of object proposals but only a few confident ones, we introduce sparsity along the detection dimension in the MIL classifier, increasing performance by emphasizing correct classification of confident proposals.

We evaluate our WS-DETR's performance on a variety of rigorous and challenging WSOD domains. For large-scale novel object detection, we utilize the Few-Shot Object Detection (FSOD) dataset~\cite{FZT20}, which has 4\(\times\) more source classes and 2\(\times\) more novel target classes than the previous largest datasets used for WSOD~\cite{RDS+15,UPF18,ZWP+20} and uses a semantic split to maximize novelty. We achieve state-of-the-art performance compared to~\cite{ZWP+20}, whose class-agnostic transfer setting is closely related to ours. Our WS-DETR is effective at WSOD training for fine-grained domains -- a practical scenario where collecting bounding box labels is difficult~\cite{HAS+18}. We evaluate fine-grained WSOD performance on the FGVC-Aircraft~\cite{MKR+13} and iNaturalist~\cite{HAS+18} datasets and demonstrate state-of-the-art results on up to 2,854 classes. Finally, our scaling study reveals that class quantity is more important than image quantity for WSOD pretraining, which may inform future dataset construction.

\section{Related Work}
\label{sec:relatedwork}

\textbf{WSOD and MIL.} %WSOD is a difficult task which has been extensively studied in the literature. 
WSOD has been formulated as an MIL problem~\cite{DLL97,ML98} since well before the deep learning era~\cite{GBR+08}. Each image is considered as a bag and the object proposals as instances, and a bag is positive for a certain class if it contains at least one instance of that class. The model is provided with candidate proposals, typically via a low-precision high-recall heuristic~\cite{USG+13,ZD14}, and learns to identify the most accurate proposals by minimizing the classification error against the bag labels.

The weakly supervised deep detection network (WSDDN)~\cite{BV16} is a particularly effective method which integrates MIL into an end-to-end learning framework, jointly optimizing a classification layer and a detection layer. Subsequent works have improved WSDDN via self-training and box refinement~\cite{JWJ+17,DSP+17,TWB+17,ZBD+18,ZFX+18,TWW+18a,ZLF+19,ZYF19,KYA19,RYY+20}, spatial relations~\cite{KOC+16,CFJ+20,TWB+20}, optimization methods~\cite{WLK+19}, residual backbones~\cite{SJW+20}, and segmentation~\cite{GLG+19,SJW+19,LKS+19}. The MIL classifier in our WS-DETR utilizes a WSDDN-like approach, but we propose using the class-agnostic model's objectness scores to directly compute the joint object and class probabilities rather than learning a detection layer.

Notable WSOD architectures not extending WSDDN include~\cite{HGT+14, AJK+19, SJC+20, AJK+21}. Self-attention methods have also been proposed~\cite{ZWK+18,CS19,YLD19,HZB+20,XLZ+21,MZT+21,PWZ+21,WWL+21,PGL+21, GWP+21}, though they focus on attention maps rather than Transformer training.

\textbf{WSOD with Knowledge Transfer.} Since initial box quality is critical to WSOD performance~\cite{HBD+15}, using a heuristic to generate proposals is often insufficient and impractical. Several strategies transfer knowledge from fully or partially annotated source datasets whose classes may not coincide with the target objects, including object count annotations~\cite{GLY+18}, semantic relationships~\cite{UPF18,TWW+18b,CDZ+21}, segmentation~\cite{SCF17,HZF+21}, objectness scores~\cite{DAF12,UPF18,LZH+19,ZWP+20}, box regression~\cite{LKC18,DHG+21}, and appearance transfer~\cite{RW15,IFY+18,RSA+20,CDZ+21}. This setting is also referred to as domain adaptation, transfer learning, or mixed supervision object detection.

Our goal is to enable streamlined knowledge transfer from a large source dataset to a variety of target domains, so we leverage the objectness scores setting by using the proposals and scores of a pretrained class-agnostic model in WSOD. We primarily compare our method to Zhong \etal \cite{ZWP+20}, the state-of-the-art Faster R-CNN approach in this setting. Besides the architecture, another key difference is that \cite{ZWP+20} allows multiple refinements over the source dataset, while our method only needs one pretraining iteration.

\textbf{Detection Transformer (DETR).} %Transformers~\cite{VSP+17}, a self-attention-based model which learns global dependencies, have introduced a new paradigm in natural language understanding and computer vision. 
In comparison to CNNs, Transformer-based methods \cite{VSP+17} require more data and training time, but excel in the large-data regime and are particularly effective at transfer learning~\cite{DBK+21}. DETR~\cite{CMS+20} introduces a Transformer framework for object detection problems. In the DETR pipeline, image features are extracted with a ResNet~\cite{HZR+15} backbone before being passed into a Transformer encoder. The decoder takes as input a number of learned object queries and attends to the encoder output, from which a feedforward network produces box and class predictions. Unlike Faster R-CNN~\cite{RHG+15}, DETR is trained with a set prediction loss and does not require non-maximum suppression, spatial anchors, or RoI pooling.

The community has made significant progress improving DETR with faster convergence~\cite{ZSL+21,GZW+21,DCY+21} and pretraining tasks~\cite{DCL+21,BWK+21}. We extend Deformable DETR~\cite{ZSL+21}, which uses an efficient attention mechanism for 10\(\times\) faster training, and DETReg~\cite{BWK+21}, which uses unsupervised pretraining to improve downstream localization performance.

To the best of our knowledge, we propose the first MIL-based DETR variant amenable to WSOD tasks. Notably, Chen \textit{et al.}~\cite{CYZ+21} extend DETR to weak point-based annotations, which is a different setting that does not involve MIL.

%\textbf{Sparsity for Object Detection.} Object detection models such as Faster R-CNN~\cite{RHG+15} utilize a dense set of region proposals, requiring a number of hand-designed components and increasing training sensitivity~\cite{SZJ+21}. Sparse R-CNN~\cite{SZJ+21} learns a small set of object proposals, avoiding the massive number of candidates generated by a region proposal network. DETR~\cite{CMS+20} learns a sparse set of object queries, though each query interacts with the global image features; Deformable DETR's~\cite{ZSL+21} efficient attention mechanism remedies this, attending to a small set of keypoints rather than the entire feature map.

%This issue is exacerbated in WSOD; with only a few true boxes out of hundreds or thousands of proposals, the MIL training process is very noisy. In addition, without box- or count-based annotations~\cite{GLG+19}, MIL does not have a bias towards the number of objects contained in a positive bag. While sparse MIL has been studied~\cite{BM07,ZNG+19}, we are to the best of our knowledge the first to introduce sparsity into an MIL-based WSOD model. \tyler{Do we need to include the sparsity section?}\neel{I don't think it is super important as long as no othre paper has used it for WSOD before.}

\section{Weakly Supervised Detection Transformer}

We propose the Weakly Supervised Detection Transformer (WS-DETR), which integrates DETR with MIL for scaling WSOD finetuning on novel objects. %WS-DETR inherits its most appealing strengths from DETR, which are particularly suited to WSOD tasks. 
In contrast to prior work requiring multiple rounds of box refinement~\cite{TWB+17} or pseudo ground truth mining~\cite{ZWP+20}, WS-DETR simplifies the process, requiring only pretraining and a single round of MIL training. Like other Transformer-based methods, WS-DETR is particularly scalable to large pretraining datasets. For WSOD applications, this enables a more accurate understanding of objectness, resulting in higher performance during knowledge transfer to novel objects. %On the other hand, the model efficacy may suffer without a sufficiently large pretraining dataset in terms of both images and classes.

\subsection{Class-Agnostic DETR}
During pretraining, we utilize a class-agnostic DETR trained on the binary object labels of the source dataset to predict bounding box proposals and objectness confidence scores for use during WSOD finetuning. The class-agnostic DETR model extends Deformable DETR~\cite{ZSL+21}, a variant of the vanilla DETR~\cite{CMS+20} with multi-scale features and improved training efficiency, and predicts a fixed-set size of \(N\) object proposals. We use \(N=300\), the default for Deformable DETR. 
% where \(N\) is set to be greater than the typical number of objects per image; we use \(N=300\), the default for Deformable DETR. 
After the Transformer features are computed, a 3-layer network with ReLU returns proposal coordinates \(\{\vb{p}_j\}_{j=1}^N\) and a fully-connected layer returns classification logits \(\{s_j\}_{j=1}^N\) interpreted as objectness scores. %The closer the source dataset is to the target dataset in terms of object similarity and environment, the more accurate these proposals and scores will be; quantifying this relationship precisely is an open challenge.

During weakly supervised finetuning, the fully-connected layer is dropped in favor of two \(C\)-class layers for classification and detection, respectively.

Optionally, we can additionally train a fully supervised class-aware DETR (\textit{i.e.,} using supervised class labels instead of binary object labels) on the source dataset. If so, the class-agnostic model is used for object proposals and scores, while the class-aware model is used for initializing the weakly supervised branch. This strategy begets a performance boost (discussed in Section \ref{sec:addtl}) and is convenient as many pretrained models are class-aware to begin with. On the other hand, using a single class-agnostic model as both proposal generator and initialization halves computation.

A key difference between our WS-DETR and previous WSOD models based on Faster R-CNN~\cite{RHG+15,ZWP+20} is that DETR learns end-to-end using a positional embedding -- also called object queries -- instead of using RoI pooling. Thus, we freeze both the object queries and box prediction head of the class-agnostic model when finetuning the pretrained DETR. If using a class-aware checkpoint, the object queries are set equal to those of the class-agnostic model; hence, the class-aware Transformer decoder attends to the same locations as its class-agnostic counterpart.

\subsection{MIL Classifier}
\label{sec:mil}
As illustrated in Figure~\ref{fig:arch}, the MIL classifier
% , based on WSDDN~\cite{BV16}, 
receives the classification logits \(\mathbf{C}\in \mathbb{R}^{N\times C}\) and detection logits \(\mathbf{D}\in \mathbb{R}^{N\times C}\) and converts them to an image-level classification prediction. The classification logits are softmaxed over the class dimension (columns), while the detection logits are softmaxed over the detection dimension (rows). Let $\sigma$ denote the softmax operation for $\vb{z}\in \mathbb{R}^N$:
\begin{equation}
    \sigma(\vb{z})_i = \frac{\exp{z_i}}{\sum_{j=1}^N \exp(z_j)}.
\end{equation}
We define the class-wise and detection-wise softmaxes as $\sigma_{ij}^c(\mathbf{A})=\sigma((\mathbf{A}^\top)_j)_i$ and $\sigma_{ij}^d(\mathbf{A})=\sigma(\mathbf{A}_i)_j$ where $\mathbf{A}_i$ is the $i^{th}$ row of $\mathbf{A}$. The softmaxed matrices in the MIL classifier are $\sigma^c(\mathbf{C})$ and $\sigma^d(\mathbf{D})$; they are then multiplied element-wise and summed over the detection dimension to obtain the image-level classification predictions \(\{\hat{y}_j\}_{j=1}^C\):
\begin{equation}\label{eq:im1}
    \hat{y}_j = \sum_{i=1}^N \sigma_{ij}^c(\mathbf{C}) \sigma_{ij}^d(\mathbf{D}).
\end{equation}

Since the rows of \(\sigma^c(\mathbf{C})\) and the columns of \(\sigma^d(\mathbf{D})\) are each non-negative and sum to one, we have \(\hat{y}_i \in (0,1)\) for all \(i\). Then, the image-level labels \(\{y_j\}_{j=1}^C\) are used to compute the negative log-likelihood loss:
\begin{equation}
\mathcal{L}_{\text{MIL}} = - \frac{1}{C} \sum_{j=1}^C y_j \log(\hat{y}_j) + (1-y_j) \log (1-\hat{y}_j).
\end{equation}

%Different from~\cite{ZWP+20}, we do not use a \(\beta\)-scaled version of the post-sigmoid detection logits in the softmax operation; instead, we use the raw detection logits as is standard in WSDDN and other works~\cite{BV16,TWB+17,HZB+20}. We observed this technique to be more consistent with minimal performance impact, and it reduces the number of hyperparameters of the algorithm.

The state-of-the-art method for knowledge transfer from a pretrained class-agnostic model is the objectness regularization technique of~\cite{ZWP+20}, which uses the class-agnostic model's objectness scores to guide the training of the detection branch. Let \(S(x)= 1/(1+e^{-x})\) denote the sigmoid operation for \(x\in\mathbb{R}\), then
\begin{equation}
\mathcal{L}_{\text{obj}} = \frac{1}{N} \sum_{i=1}^N \big(\max_{1\leq j \leq C} S(\mathbf{D}_{ij}) - S(s_i)\big)^2.
\end{equation}

%Applying the regularization with coefficient \(\lambda\), the model loss is
%\begin{equation}
%    \mathcal{L} = %\mathcal{L}_{\text{MIL}} + \lambda %\mathcal{L}_{\text{obj}}.
%\end{equation}
Hence, the model loss is $\mathcal{L} = \mathcal{L}_{\text{MIL}} + \lambda \mathcal{L}_{\text{obj}}$ for a coefficient $\lambda$. During inference, WS-DETR returns box \(\vb{p}_i\) with class prediction and confidence determined by $\argmax_{1\leq j \leq C} \sigma^c_{ij}(\mathbf{C})\sigma^d_{ij}(\mathbf{D})$.
%\begin{equation}
%    \argmax_{1\leq i \leq N} %\sigma^c_{ij}(\mathbf{C})\sigma^d_{ij}(\%mathbf{D}).
%\end{equation}

\subsection{Joint Probability Estimation}
\label{sec:likelihood}

We show in Section \ref{sec:fgvc} that the objectness regularization technique of \cite{ZWP+20} is insufficient for general WSOD, as it can suffer from the common MIL weakness of overfitting to distinctive classification features~\cite{TWB+17}. To rectify this and more effectively utilize the pretrained knowledge from DETR, we propose a formulation for the MIL classifier based on the joint object and class probabilities for each proposal~\cite{RF17}. For a given proposal $i$, let $c_i = \max_{1\leq j \leq C} \sigma_{ij}^c(\mathbf{C})$ and $d_i = \max_{1\leq j \leq C} S(\mathbf{D}_{ij})$ be its maximum classification and detection probabilities, respectively.

There are two scenarios which can cause this overfitting problem. First, if $c_i$ is high for a particular class $j$ but $\mathbf{D}_{ij}$ is low, the model may take a penalty in $\mathcal{L}_{\text{obj}}$ to increase $\mathbf{D}_{ij}$ and easily minimize $\mathcal{L}_{\text{MIL}}$ for the image. However, this can be avoided by increasing $\lambda$. The more likely explanation is based on a weakness of the objectness regularizer: for a given proposal $i$, the regularizer only cares about the value of $d_i$ and not whether its position in the row actually lines up with $c_i$ -- that is, whether
\begin{equation}
\argmax_{1\leq j \leq C} \sigma_{ij}^c(\mathbf{C}) = \argmax_{1\leq j \leq C} S(\mathbf{D}_{ij}).
\end{equation}
If these values are mismatched, this failure case would result in low confidences for every proposal and essentially sort them by $c_i$, causing overfitting. Indeed we observe that when using our WS-DETR with objectness regularization, the final confidences are typically 0.01 or lower.

We desire the overall probability of these distinctive features to be diminished by a low objectness probability, since the pretrained model should recognize that the feature does not represent an entire object. Thus, we compute the probability $\mathbb{P}[i^{th} \text{ proposal is an object and instance of class } j] = \sigma_{ij}^c(\mathbf{C})S(s_i)$. With this formulation, it is required that a certain proposal should have both a nontrivial classification and objectness probability for it to be included in the final prediction. Using normalized probabilities via softmax \cite{BV16}, we obtain the new image-level classification prediction
\begin{equation}\label{eq:im2}
    \hat{y}_j = \sum_{i=1}^N \sigma_{ij}^c(\mathbf{C})\sigma(\vb{s})_i.
\end{equation}

Note that our technique is mutually exclusive with objectness regularization. We show in Section \ref{sec:fgvc} that our modification successfully prevents overfitting to distinctive classification features. Further, our technique simplifies the network as we are essentially using the objectness confidences from the pretrained DETR in place of a learnable detection branch in the MIL classifier, and it improves convergence by minimizing $\mathcal{L}_{\text{MIL}}$ without $\mathcal{L}_{\text{obj}}$.

\subsection{Sparsity in the MIL Classifier}
\label{sec:sparsity}
The objectness knowledge present in the pretrained DETR can also be leveraged to reduce noise during multiple instance learning -- while there are a fixed number of \(N=300\) proposals, the model typically only detects a few with high objectness scores. To focus more on these confident proposals, we propose utilizing sparsity along the detection dimension of the MIL classifier.

To do so, we apply the sparsemax function~\cite{MA16} instead of the softmax function along the detection dimension in the MIL classifier; this operation zeros out some low-confidence boxes, increasing emphasis on correct classification of likely proposals. Specifically, sparsemax returns the Euclidean projection of a vector \(\vb{z}\in \mathbb{R}^N\) onto the \((N-1)\)-dimensional probability simplex \(\Delta^{N-1} = \{\vb{p}\in\mathbb{R}^N:\vb{1}^\top\vb{p}=1, \vb{p}\geq \vb{0}\}\):
\begin{equation}
    \text{sparsemax}(\vb{z}) = \argmin_{\vb{p}\in\Delta^{N-1}} \|\vb{p}-\vb{z}\|_2^2.
\end{equation}
We then substitute $\text{sparsemax}(\mathbf{D}_i)_j$ for \(\sigma^d_{ij}(\mathbf{D})\) in Equation \ref{eq:im1} and $\text{sparsemax}(\vb{s})_i$ for $\sigma(\vb{s})_i$ in Equation \ref{eq:im2}.

\begin{table}[t]
    \footnotesize
    \centering
    \caption{Class-agnostic performance of Faster R-CNN (used by \cite{ZWP+20}) and DETR methods trained on FSOD-800 and evaluated on each FSOD-200 test split, ignoring classes. We use the codebase of \cite{ZWP+20},  which does not report precision for this task.}
    \begin{tabular}{|l c c c|}
    \hline
    Method & mAP & AP50 & mAR \\
    \hline\hline
    Zhong \etal~\cite{ZWP+20} &$-$ &$-$ & $50.5 \pm 2.1$\\
    Class-Aware DETR & $18.4 \pm 1.0$ & $26.9\pm 0.94$ & $62.3 \pm 3.0$\\
    Class-Agnostic DETR & $\vb{30.6 \pm 1.6}$ & $\vb{43.0 \pm 1.6}$ & $\vb{65.5 \pm 3.2}$\\
    \hline
    \end{tabular}
    \label{tab:fsod_agnostic}
\end{table}
While there are many sparsity techniques, we choose sparsemax because of its theoretical justification, its ease-of-use with no hyperparameter tuning, and its successful application in previous MIL architectures~\cite{ZNG+19} (though not in WSOD). The structure of the loss function also makes it particularly well-suited to the MIL problem. 
In a traditional classification setting, it is possible to get a \(\log(0)\) in the loss function because sparsemax can send labels to a probability of exactly zero -- Martins and Astudillo~\cite{MA16} define a new sparsemax loss to deal with this issue. But because we multiply \(\text{sparsemax}(\mathbf{D})\) element-wise with \(\sigma^c(\mathbf{C})\) whose entries are $>0$, there is some element $>0$ in each column of the product. Thus, we still have \(\hat{y}_i\in (0,1)\) for all \(i\), and we can still apply the negative log-likelihood loss. 

%Intriguingly, our experiments in Section \ref{sec:experiments} show that, while sparsity by itself still suffers from the class-overfitting problem, combining it with our joint probability technique from Section \ref{sec:likelihood} results in consistently increased performance.

\section{Experiments}
\label{sec:experiments}

%We demonstrate the effectiveness of our WS-DETR approach in three distinct scenarios. First, we . Our model 

\subsection{Large-Scale Novel Object Detection}
\label{sec:fsod}

To evaluate the performance of our WS-DETR on highly novel objects, we utilize the the Few-Shot Object Detection (FSOD) dataset~\cite{FZT20}, designed to test the generalization performance of few-shot learning models on novel objects in a diverse setting. Fan \etal~\cite{FZT20} constructed the dataset from existing large-scale supervised datasets ILSVRC~\cite{RDS+15} and Open Images~\cite{KRA+20}, and merged their semantic trees pursuant to Open Images superclasses. The FSOD dataset comprises 1000 classes, of which 800 are reserved for training and 200 for testing -- we call these datasets FSOD-800 and FSOD-200. This train/test split is generated such that the test classes have the largest distance from existing training categories in the semantic tree, enabling a challenging setting for generalization to truly novel objects.

\begin{table}[t]
    \centering
    \caption{WSOD performance on FSOD-200 splits with FSOD-800 pretraining. Our WS-DETR is initialized with class-agnostic proposal generator and class-aware weights. The supervised DETR is finetuned from the class-aware FSOD-800 checkpoint.}
    \footnotesize
    \begin{tabular}{|l c c c|}
    \hline
    Method & mAP & AP50 & mAR\\
    \hline\hline
    Zhong \etal~\cite{ZWP+20} & $20.6\pm 0.76$ & $32.7 \pm 2.0$  & $34.4 \pm 0.43$\\
    WS-DETR Base & $13.9\pm 1.6$ & $20.0 \pm 1.9$ & $60.1 \pm 2.4$\\
    WS-DETR Sparse & $28.5 \pm 0.86$ & $\vb{38.5 \pm 0.63}$ & $\bm{68.0 \pm 4.3}$\\
    WS-DETR Joint & $28.6 \pm 0.43$ & $37.8 \pm 0.87$ & $65.3 \pm 1.5$\\
    WS-DETR Full & $\vb{28.6 \pm 0.25}$ & $38.2 \pm 1.1$ & $67.4 \pm 3.9$\\
    \hline\hline
    Supervised DETR & $47.7 \pm 1.3$ & $64.0 \pm 1.0$ & $76.3 \pm 1.2$\\
    \hline
    \end{tabular}
    \label{tab:fsod}
\end{table}

In contrast to few-shot object detection, WSOD requires a target dataset of novel objects for model finetuning. Thus, we utilize FSOD-800 as a source dataset for pretraining, and we create three random train/test splits of FSOD-200 for training and evaluation on novel objects, which will be released for reproducibility. We report the mean and 95\% confidence interval of the metrics against each split based on a \(t\)-distribution with two degrees of freedom. FSOD-800 has 52,350 images with 147,489 boxes, while the FSOD-200 splits each have 11,322 training images with between 28,008 and 28,399 boxes and 2,830 testing images with between 6,703 and 7,094 boxes. This setting has \(4\times\) the source classes and \(2\times\) the target classes than the largest datasets used for WSOD previously~\cite{RDS+15,UPF18,ZWP+20}.

While some WSOD methods use the Correct Localization (CorLoc) metric~\cite{DAF10} for evaluating localization accuracy, this metric is too lenient as it only requires localizing a single object per image. Hence, we instead use the mean average recall (mAR) at 100 detections per image for comparing class-agnostic proposal quality, though we also report mean average precision (mAP) and AP50 for comparison. In Table \ref{tab:fsod_agnostic} we compare the performance of the class-agnostic and class-aware DETR versus the class-agnostic Faster R-CNN of~\cite{ZWP+20} trained on FSOD-800 and evaluated on each FSOD-200 test split. For the class-aware DETR, we ignore the class predictions and evaluate the boxes only. Both DETR variants outperform the Faster R-CNN, and the class-agnostic DETR and class-aware DETRs have similar recall, though the class-agnostic DETR has much better precision. We show in Section \ref{sec:addtl} that this precision improvement translates to superior WSOD performance, justifying the extra pretraining of a class-agnostic model.

\begin{figure}\centering
\subfloat[WS-DETR Base]{
\begin{tabular}[b]{@{}c@{}}
\includegraphics[width=.22\linewidth]{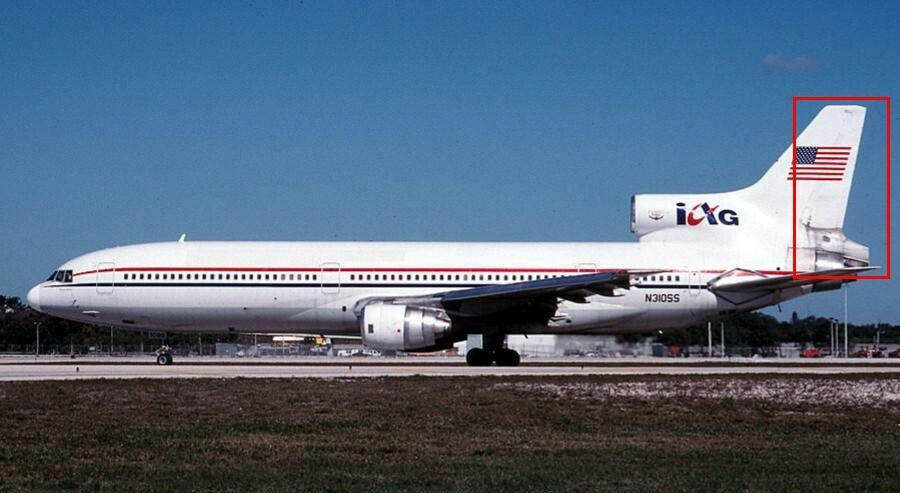}%
\includegraphics[width=.22\linewidth]{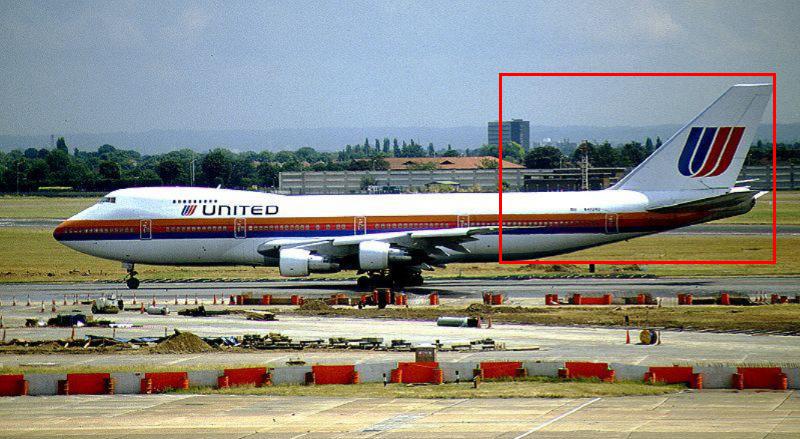}\\[-3pt]
\includegraphics[width=.22\linewidth]{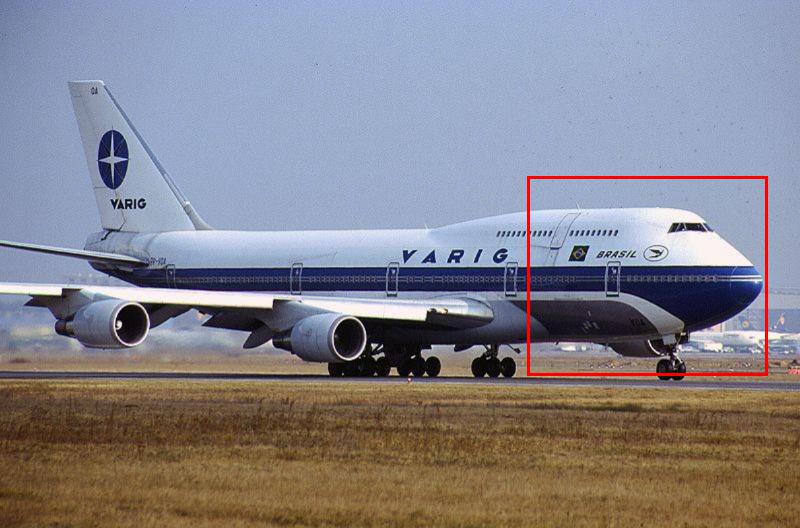}%
\includegraphics[width=.22\linewidth]{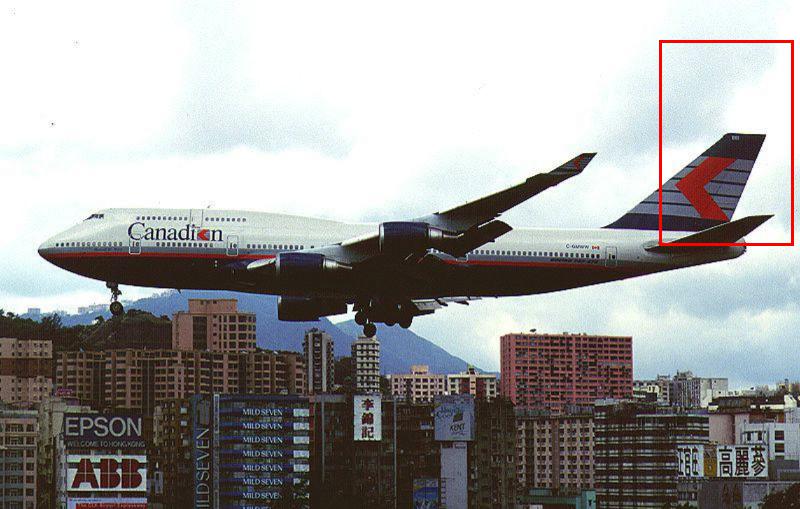}
\end{tabular}}\quad
\subfloat[WS-DETR Sparse]{
\begin{tabular}[b]{@{}c@{}}
\includegraphics[width=.22\linewidth]{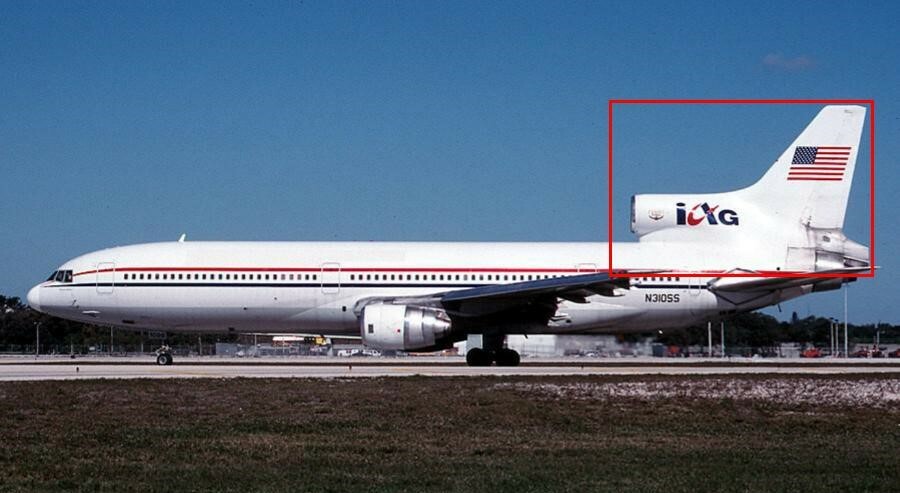}%
\includegraphics[width=.22\linewidth]{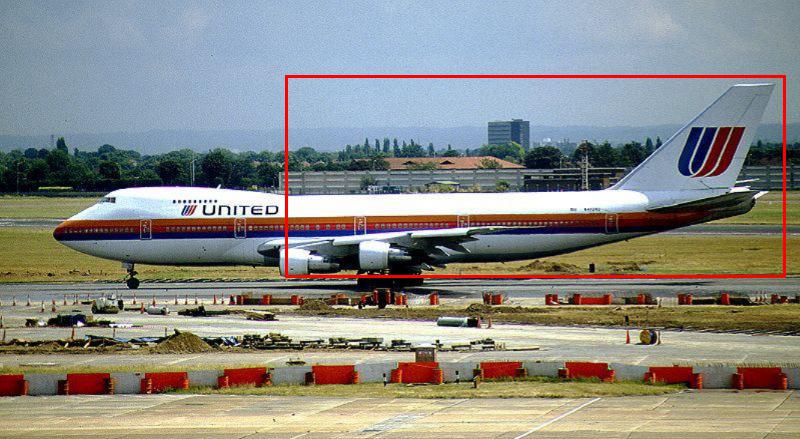}\\[-3pt]
\includegraphics[width=.22\linewidth]{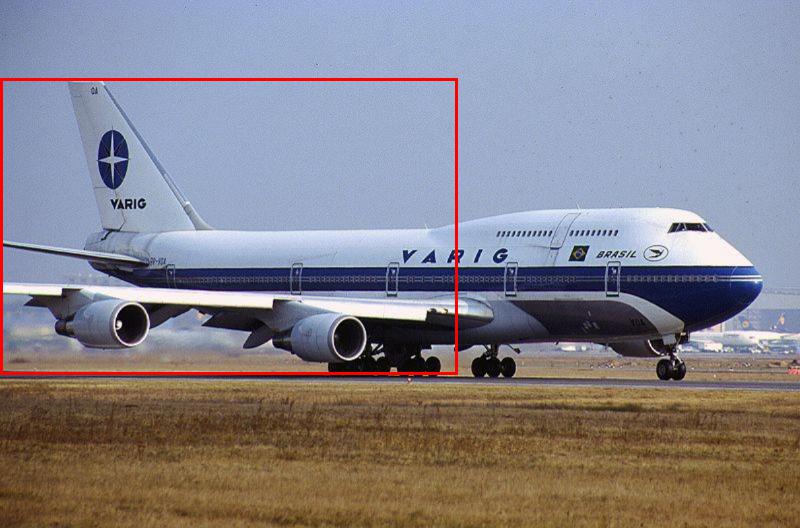}%
\includegraphics[width=.22\linewidth]{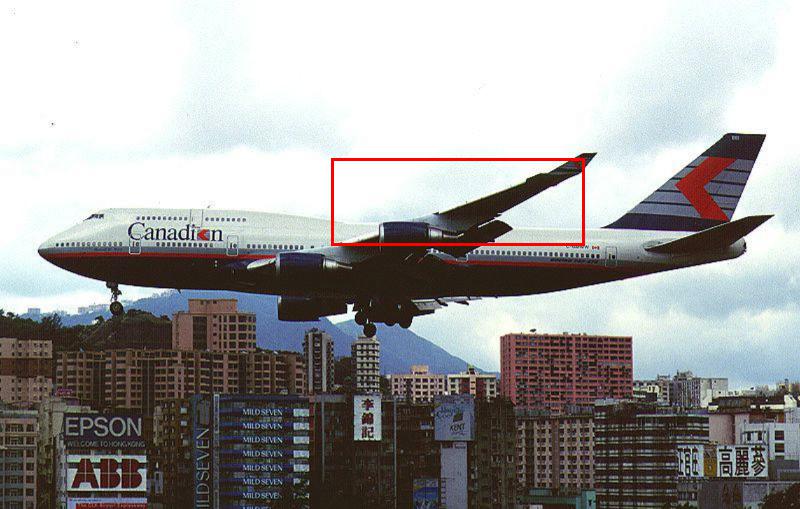}
\end{tabular}}\\
\subfloat[WS-DETR Joint]{
\begin{tabular}[b]{@{}c@{}}
\includegraphics[width=.22\linewidth]{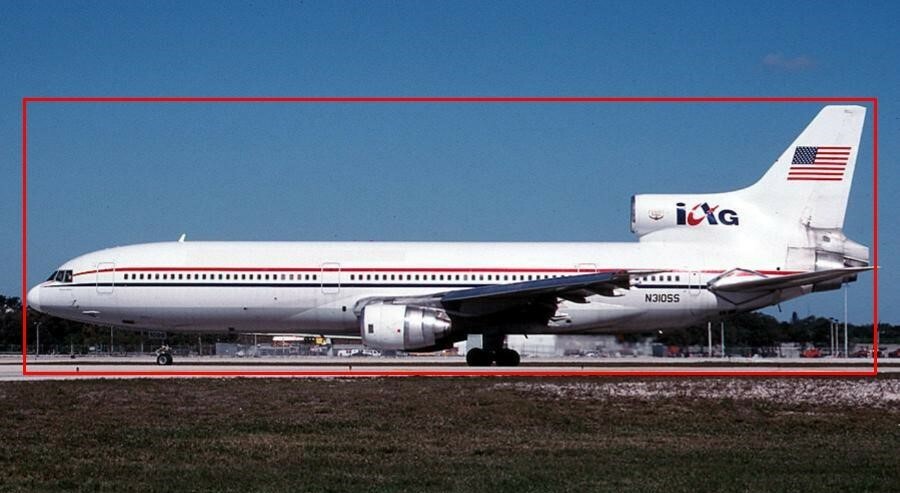}%
\includegraphics[width=.22\linewidth]{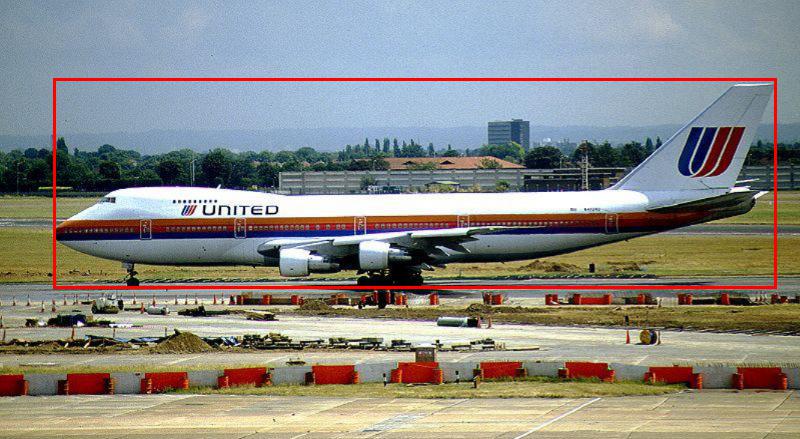}\\[-3pt]
\includegraphics[width=.22\linewidth]{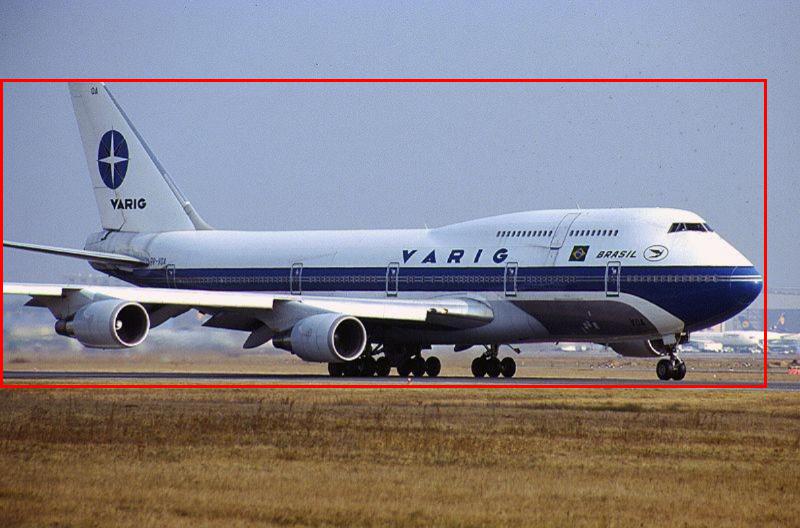}%
\includegraphics[width=.22\linewidth]{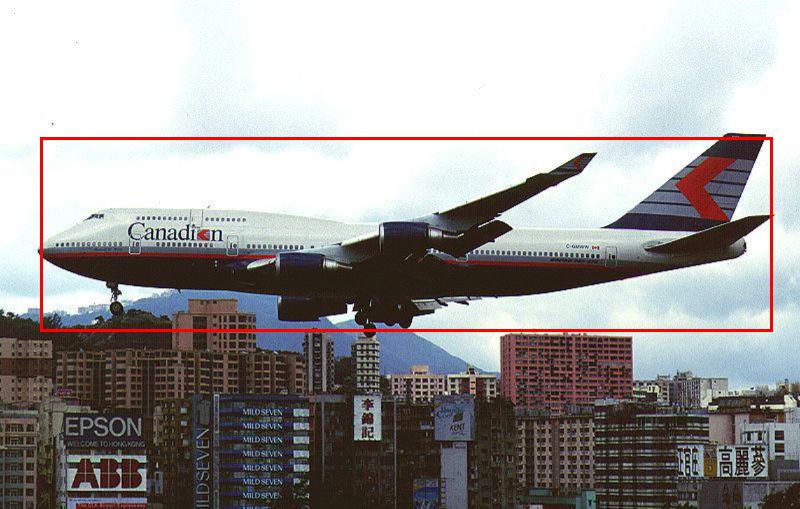}
\end{tabular}}\quad
\subfloat[WS-DETR Full]{
\begin{tabular}[b]{@{}c@{}}
\includegraphics[width=.22\linewidth]{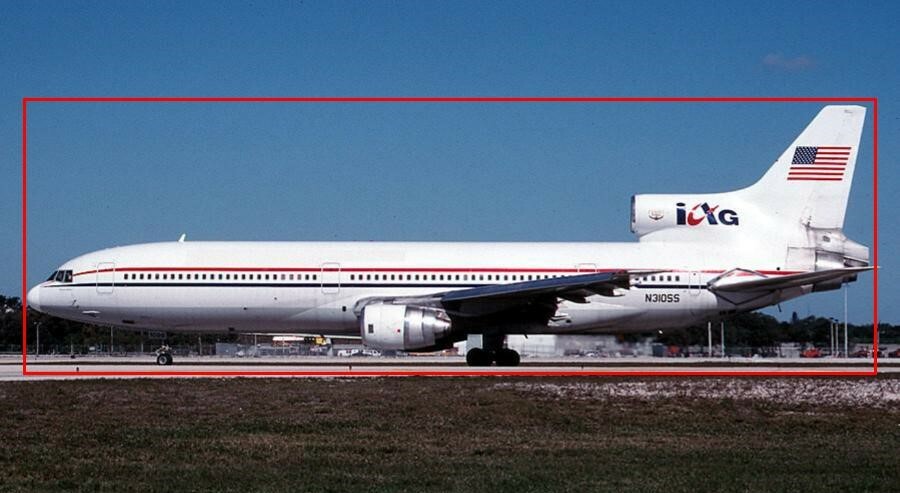}%
\includegraphics[width=.22\linewidth]{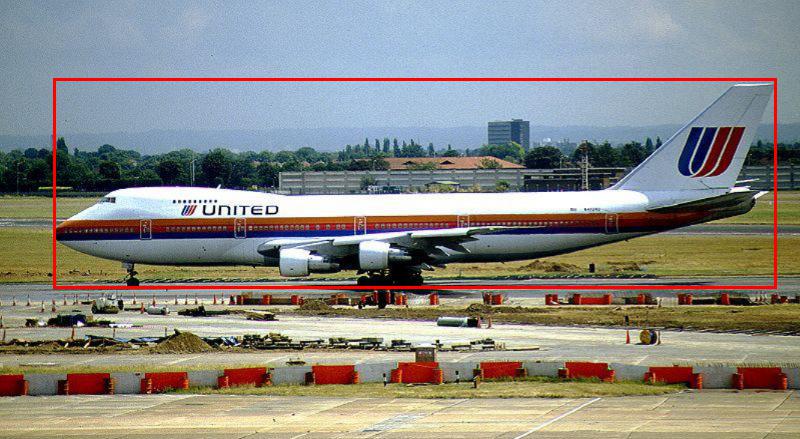}\\[-3pt]
\includegraphics[width=.22\linewidth]{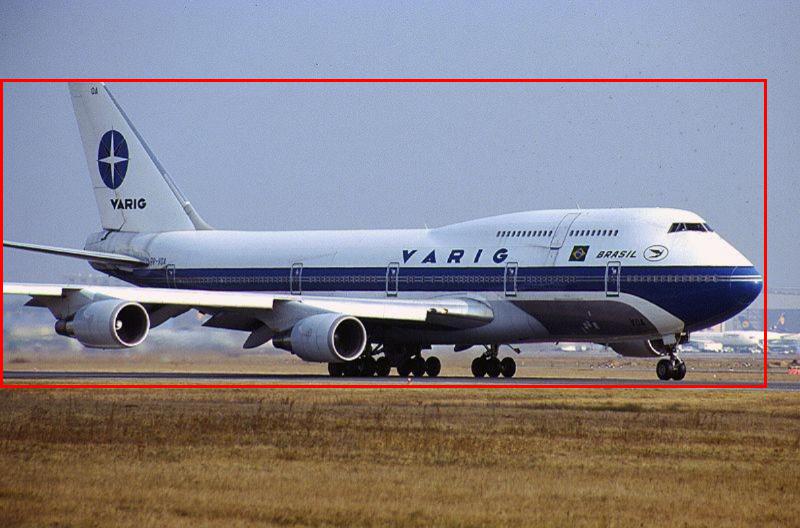}%
\includegraphics[width=.22\linewidth]{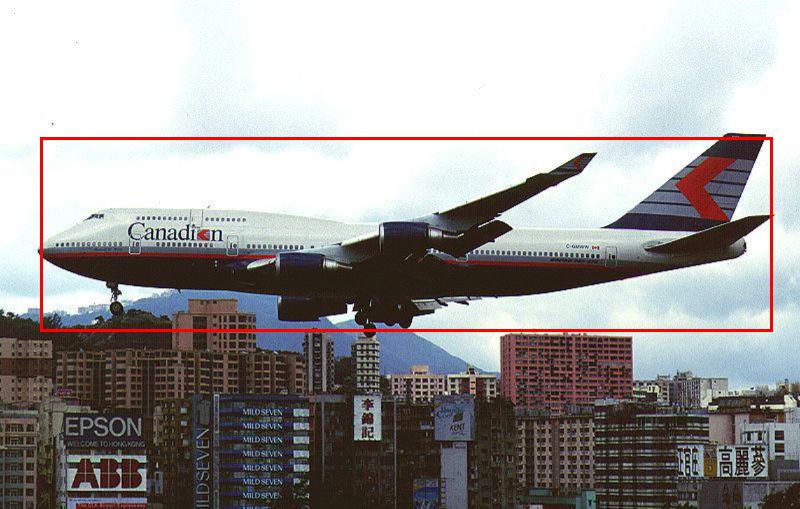}
\end{tabular}}
\caption{Visualization of how our joint probability technique prevents overfitting to distinctive classification features on the FGVC-Aircraft dataset. Best viewed electronically and zoomed in. Models (c) and (d) both use our technique. The plotted box is the highest confidence detection.}
\label{fig:fgvc_img}
\end{figure}

For brevity, we introduce short names for each permutation of WS-DETR with our techniques from Sections \ref{sec:likelihood} and \ref{sec:sparsity}. ``Base'' refers to our model with the objectness regularization of \cite{ZWP+20}; ``Sparse'' refers to ours with sparsity and objectness regularization; ``Joint'' refers to ours with joint probability estimation only; and ``Full'' refers to ours with sparsity and joint probability estimation.

We then train our WS-DETR with class-agnostic proposal generator and class-aware weights initialization on each FSOD-200 split. In Table \ref{tab:fsod}, we detail the performance of our model against the state-of-the-art baseline of \cite{ZWP+20} and a supervised DETR upper bound. The addition of either our joint probability technique or sparsity boosts mAP by nearly 15 points over WS-DETR Base, achieving a new state-of-the-art performance by 8 mAP. This suggests that, for WS-DETR, regularization by itself is insufficient for transferring objectness knowledge learned from the source dataset to the downstream WSOD task. In particular, as we show in the next section, our joint probability technique is critical to prevent overfitting to distinctive classification features. Additionally, while the method of \cite{ZWP+20} loses 15 mAR during weakly supervised training, our WS-DETR gains 2.5 mAR relative to the class-agnostic pretrained model.

\subsection{Joint Probability Prevents Overfitting}
\label{sec:fgvc}

\begin{table}[t]
    \centering
    \caption{WSOD performance on the FGVC-Aircraft dataset with FSOD-800 pretraining. The models using our joint probability technique achieve near-supervised performance, while the objectness regularization methods underperform due to overfitting to distinctive classification features.}
    \footnotesize
    \begin{tabular}{|l c c c|}
    \hline
    Method & mAP & AP50 & mAR \\
    \hline\hline
    Zhong \etal~\cite{ZWP+20} & 14.8 & 28.7 & 30.5\\
    WS-DETR Base & 5.2 & 8.5 & 63.4\\
    WS-DETR Sparse & 50.6 & 57.4 & 93.2\\
    WS-DETR Joint & 77.7 & 83.6 & 93.4 \\
    WS-DETR Full & $\vb{79.1}$ & $\vb{85.0}$ & $\vb{94.2}$ \\
    \hline\hline
    Supervised DETR & 87.1 & 88.7 & 97.9 \\
    \hline
    \end{tabular}
    \label{tab:fgvc}
\end{table}

The FGVC-Aircraft dataset \cite{MKR+13} comprises 10,000 images of 100 types of aircraft whose visual characteristics may differ only slightly between classes. It poses a fairly simple detection problem, as the target objects are large and centered. Additionally, since ``airplane'' is one of the FSOD-800 source classes, one would expect WSOD models to perform well on this task. We show that our joint probability formulation achieves this outcome, while the objectness regularization technique utilized in previous work~\cite{ZWP+20} limits detection performance. In particular, previous models overfit to distinctive classification features -- here, the nose or tail of the aircraft -- a weakness observed by~\cite{TWB+17} and whose remedy has been the subject of several WSOD studies~\cite{TWB+17,ZLF+19,HZB+20}. These solutions typically involve multiple iterative rounds of box refinement via self-training. In contrast, our method leverages the objectness knowledge from the pretrained model to identify the correct proposal without any extra computation.

In Figure \ref{fig:fgvc_img}, we visualize the advantage of our WS-DETR method and how it properly selects the bounding box covering the entire aircraft, while the objectness regularization technique overfits to distinctive features. In Table \ref{tab:fgvc}, we display the precision and recall of each model on the FGVC-Aircraft test set and demonstrate that our model achieves near-supervised level performance.

\subsection{Finetuning on 2,854 Fine-grained Classes}

A practical application for WSOD not captured by current settings is on datasets with many classes which require domain-specific knowledge for labeling. An exemplar of this variety is the iNaturalist 2017 dataset \cite{HAS+18}, a fine-grained species dataset of 500K boxes and 5,000 classes, 2,854 of which have detection annotations. In fact, Horn \etal~\cite{HAS+18} remark that the bounding box labeling was particularly difficult -- since only a domain expert can distinguish between so many different species, they asked labelers to demarcate superclasses only, then backfilled the bounding boxes with the (sometimes incorrect) image-level classification labels. The most recent iNaturalist dataset contains 10,000 species and 2.7 million images, which is nearly impossible to label for detection tasks and represents a highly practical scenario for a weakly supervised approach.

\begin{table}[t]
    \centering
    \caption{WSOD performance on the iNaturalist 2017 dataset with FSOD-800 pretraining. Our WS-DETR is initialized with class-agnostic proposal generator and class-aware weights. The supervised DETR upper bound is finetuned from the same class-aware FSOD-800 checkpoint. The method of~\cite{ZWP+20} did not converge for the subclasses task.}
    \scriptsize
    \begin{tabular}{|l c c c c c c|}
    \hline
    \multirow{2}{*}{Method} & \multicolumn{3}{c}{13 Superclasses} & \multicolumn{3}{c|}{2,854 Subclasses} \\
    & mAP & AP50 & mAR & mAP & AP50 & mAR\\
    \hline\hline
    Zhong \etal~\cite{ZWP+20} & $44.1$ & $76.7$ & 57.1 &$-$ & $-$ & $-$\\
    WS-DETR Base & $0.2$ & $0.4$ & 31.9 & 1.7 & 3.7 & 26.3\\
    WS-DETR Sparse & $\vb{61.1}$ & $\vb{79.3}$ & 83.3 & 30.4 & 38.2 & $\vb{77.6}$\\
    WS-DETR Joint & $54.8$ & $70.0$ & $\vb{84.3}$ & 22.1 & 29.8 & 75.5\\
    WS-DETR Full & $60.7$ & $78.7$ & 83.1 & $\vb{35.4}$ & $\vb{43.5}$ & 75.5\\
    \hline\hline
    Supervised DETR & $79.2$ & $93.6$ & 88.6 &51.5 & 58.8 & 85.6\\
    \hline
    \end{tabular}
    \label{tab:inat}
\end{table}
The WS-DETR model has seen different types of plants and animals during pretraining on FSOD-800, but nowhere near the granularity and diversity of iNaturalist with its thousands of leaf classes. This makes iNaturalist an interesting setting for studying knowledge transfer during WSOD training. In Table \ref{tab:inat}, we detail the performance of our WS-DETR against a state-of-the-art model~\cite{ZWP+20} on the 13 superclasses and 2,854 subclasses in the dataset. While the method of~\cite{ZWP+20} did not converge on the subclasses, our model achieves 75\% of supervised performance.

The iNaturalist experiment reveals several intriguing phenomena. First, our WS-DETR does not converge without our joint probability technique or sparsity, suggesting that our techniques improve training stability as well as performance. Second, the addition of sparsity to our joint probability technique improves results by up to 8.3 mAP, showing its versatility and effectiveness even without a fully-connected detection branch. Third, our WS-DETR outperforms the state-of-the-art by 17 mAP but less than 3 AP50; this indicates that, while Faster R-CNN is able to localize objects at a modest IoU threshold, the improved localization and knowledge transfer in our WS-DETR can have a significant impact on high-precision WSOD performance.

We note that, while WS-DETR Sparse sometimes attains the top performance, we still recommend WS-DETR Full for implementation in practice. WS-DETR Full consistently scores within 0.5 mAP of the best model, and without prior knowledge as to the distinctive features of the target dataset, WS-DETR Sparse may overfit due to its reliance on objectness regularization. Hence, as seen in FGVC-Aircraft and the iNaturalist subclasses, WS-DETR Full greatly outperforms WS-DETR Sparse in fine-grained datasets.

\subsection{COCO-60 to VOC Performance}
\begin{table}[t]
    \centering
    \caption{WSOD performance on PASCAL VOC 2007 with COCO-60 pretraining. The supervised DETR is finetuned from the same COCO-60 checkpoint. The result of Zhong \etal~\cite{ZWP+20} includes pseudo ground truth mining.\\}
    \footnotesize
    \begin{tabular}{|l c c c |}
    \hline
    Method & mAP & AP50 & mAR\\
    \hline\hline
    WSDDN~\cite{BV16} & $-$ & 34.8 & $-$\\
    CASD~\cite{HZB+20} & $-$ & 56.8 & $-$ \\
    Zhong \etal~\cite{ZWP+20} & $-$ & $\vb{59.7}$ & $-$ \\
    WS-DETR Base & 18.2 & 28.4 & 58.4\\
    WS-DETR Sparse & 24.2 & 36.5 & 57.7\\
    WS-DETR Joint & 23.4 & 33.8 & 58.4\\
    WS-DETR Full & 23.6 & 34.2 & 57.6\\
    \hline\hline
    Supervised DETR & 55.3 & 77.3 & 72.7 \\
    \hline
    \end{tabular}
    \label{tab:voc}
\end{table}
Though the presence of many classes in the source dataset is posited to be essential for effective transfer~\cite{UPF18}, previous WSOD methods are typically designed for and trained on datasets with few classes and small image sets. One such setting is PASCAL VOC~\cite{EGW+10} (20 classes), where knowledge transfer methods use COCO-60~\cite{LMB+14,LKC18} (60 classes) for pretraining with no class overlap.

As stated in Section \ref{sec:introduction}, reliance on COCO-60/VOC has drawbacks which limit the usage of previous WSOD models in practice. Yet, for completeness we have tested our method on this common test case (detailed in Table~\ref{tab:voc}). Our best approach is below the leading method~\cite{ZWP+20}. This is at odds with our performance relative to~\cite{ZWP+20} for diverse datasets with hundreds of novel objects such as FSOD and fine-grained datasets such as iNaturalist and FGVC-Aircraft. We believe this inconsistency illustrates the benefits of our method, which can leverage large-scale pretraining for weakly supervised detection on complex datasets that are common in real-world scenarios. Our results suggest this benefit is due to the scalability of our model and our novel method of finetuning an end-to-end detection model instead of a classification model such as ResNet \cite{HZR+15,ZWP+20}; however, a limitation of our model -- shared with all Transformer methods -- is its requirement of a large pretraining dataset for effective downstream performance.

Previous methods work well for the small-scale COCO-60/VOC case but not large and diverse datasets, which we believe are more common in real-world applications. Together with our scaling study in Section \ref{sec:addtl}, this shows that it is time for WSOD research to move beyond what appears to be an over-optimization to COCO-60/VOC, which is not a useful analogue for real-world datasets, and address the complex datasets that we study in our work.

\subsection{Ablation Study}
\label{sec:addtl}

\begin{figure}[t]
    \centering
    \includegraphics[scale=0.45]{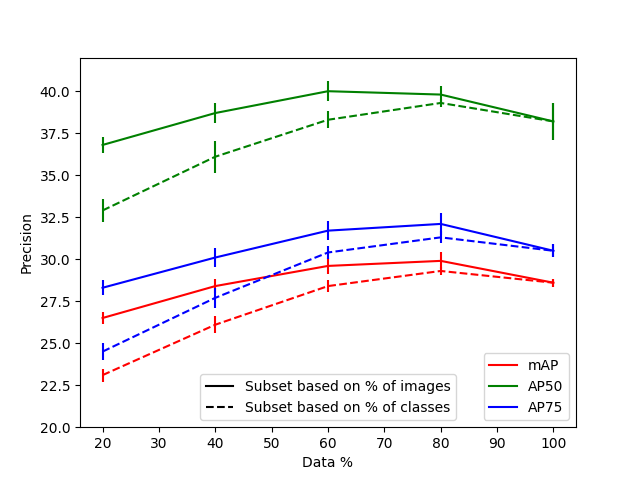}
    \caption{Scaling study of FSOD-800 pretrained WS-DETR Full with FSOD-200 WSOD. We test pretraining with a percentage of images vs. a percentage of classes, then perform WSOD training and evaluate on our held-out test set. This shows pretraining class quantity contributes more to performance than image quantity.}
    \label{fig:scaling}
\end{figure}

We perform a scaling study on FSOD-800 pretraining with WS-DETR Full and find that class quantity contributes more to downstream WSOD performance than image quantity (see Figure \ref{fig:scaling}). Group 1, the solid lines in the figure, are a random split of FSOD-800 with all classes represented. Group 2, the dashed lines in the figure, have the same number of images as Group 1 but with that same proportion of classes. This experimental setup isolates the effect of increased pretraining classes with the same number of total images. We take 3 random splits of FSOD-800 at each percentage level for each group and finetune on the 3 splits of FSOD-200. We report the mean and 95\% CI with respect to a $t$-distribution with 8 \textit{dof}. This is the first rigorous testing and proof of the hypothesis of Uijlings \etal~\cite{UPF18} that class quantity is more important than image quantity for WSOD pretraining, and it justifies our usage of FSOD-800 in place of a larger dataset with less classes such as COCO~\cite{LMB+14}. The lowest proportion of classes we test (160 classes) is still nearly 3$\times$ that of COCO-60~\cite{LMB+14,LKC18}; the performance gap at this level suggests that standard datasets used for WSOD pretraining are an order of magnitude too small.

We note that Figure \ref{fig:scaling} peaks around 80\% of data; we believe this is due to the random dropout of irrelevant classes from the pretraining dataset. A future direction is to quantify the impact of class diversity on WSOD performance; indeed, we and Uijlings \etal~\cite{UPF18} use quantity as a proxy for diversity, which has been shown in theoretical works~\cite{DHK+21} to be necessary for effective finetuning on novel classes.

\begin{table}[t]
    \centering
    \caption{WSOD performance on each FSOD-200 split with FSOD-800 pretraining. We utilize our joint probability technique and no sparsity. The class-aware DETR is pretrained on FSOD-800 with class labels, while the class-agnostic DETR is pretrained with only binary object labels.}
    \footnotesize
    \begin{tabular}{|c c c c|}
    \hline
    Proposal Generator & Weights Init. & mAP & AP50\\
    \hline\hline
    Aware & Agnostic & $18.0 \pm 1.1$ & $24.3 \pm 1.4$\\
    Aware & Aware & $22.1 \pm 1.7$ & $29.7 \pm 2.3$\\
    Agnostic & Agnostic & $27.0 \pm 1.0$ & $35.8 \pm 1.7$\\
    Agnostic & Aware & \vb{$28.6 \pm 0.43$} & \vb{$37.8 \pm 0.87$}\\
    \hline
    \end{tabular}
    \label{tab:init}
\end{table}

In our above experiments, we used a class-agnostic pretrained DETR as the proposal generator and a class-aware pretrained DETR as the weights initialization. This requires pretraining two separate DETR models, which can be computationally intensive for large source datasets. In these cases, we can halve computation by using the class-agnostic model as the weights initialization. In Table \ref{tab:init}, we detail the performance of different proposal generator and weights initialization strategies on each FSOD-200 split. The WS-DETR trained with class-agnostic proposal generator and weights initialization only loses 1.6 mAP and 2.0 AP50 compared to the best model; this suggests that while the feature space learned during the class-aware pretraining is a useful initialization, the class-agnostic model can learn most necessary features during WSOD finetuning.

\begin{figure}\centering
\begin{subfigure}[b]{0.155\textwidth}
\includegraphics[width=\textwidth]{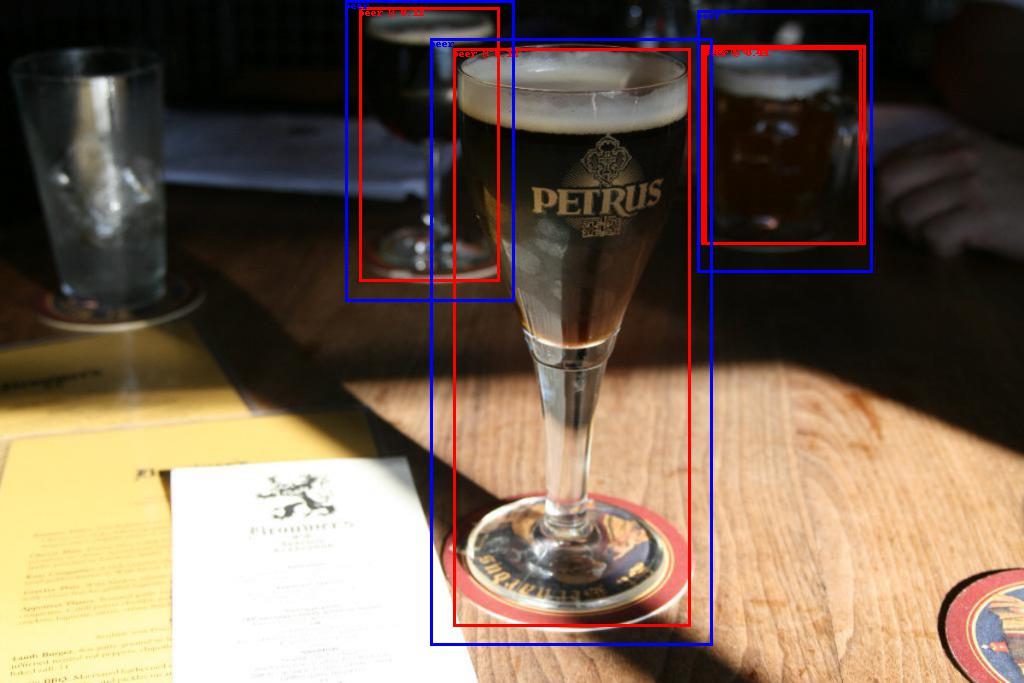}
\end{subfigure}
\begin{subfigure}[b]{0.155\textwidth}
\includegraphics[width=\textwidth]{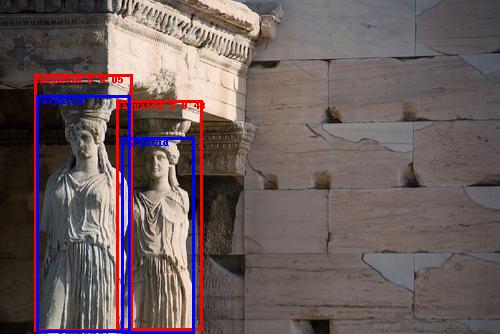}
\end{subfigure}
\begin{subfigure}[b]{0.155\textwidth}
\includegraphics[width=\textwidth]{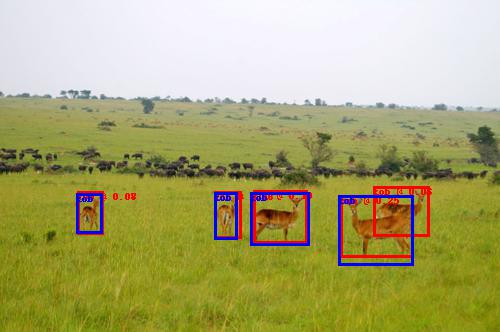}
\end{subfigure}
\begin{subfigure}[b]{0.155\textwidth}
\includegraphics[width=\textwidth]{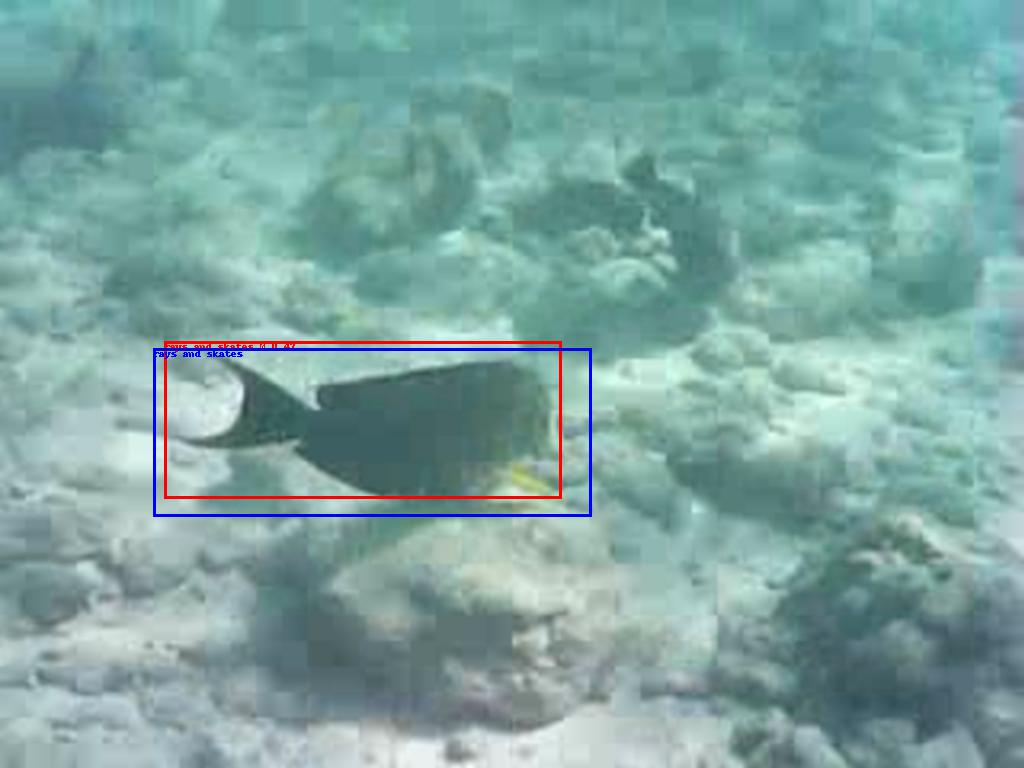}
\end{subfigure}
\begin{subfigure}[b]{0.155\textwidth}
\includegraphics[width=\textwidth]{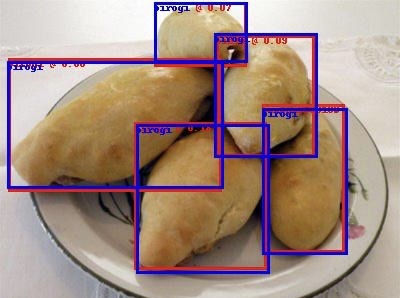}
\end{subfigure}
\begin{subfigure}[b]{0.155\textwidth}
\includegraphics[width=\textwidth]{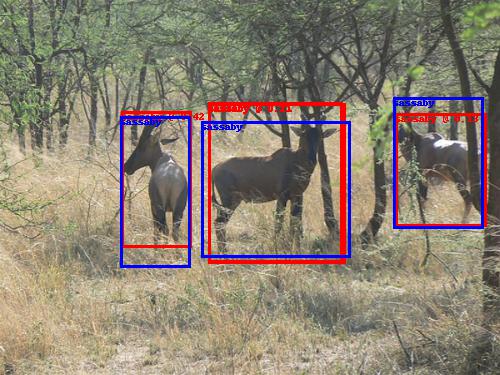}
\end{subfigure}
\begin{subfigure}[b]{0.155\textwidth}
\includegraphics[width=\textwidth]{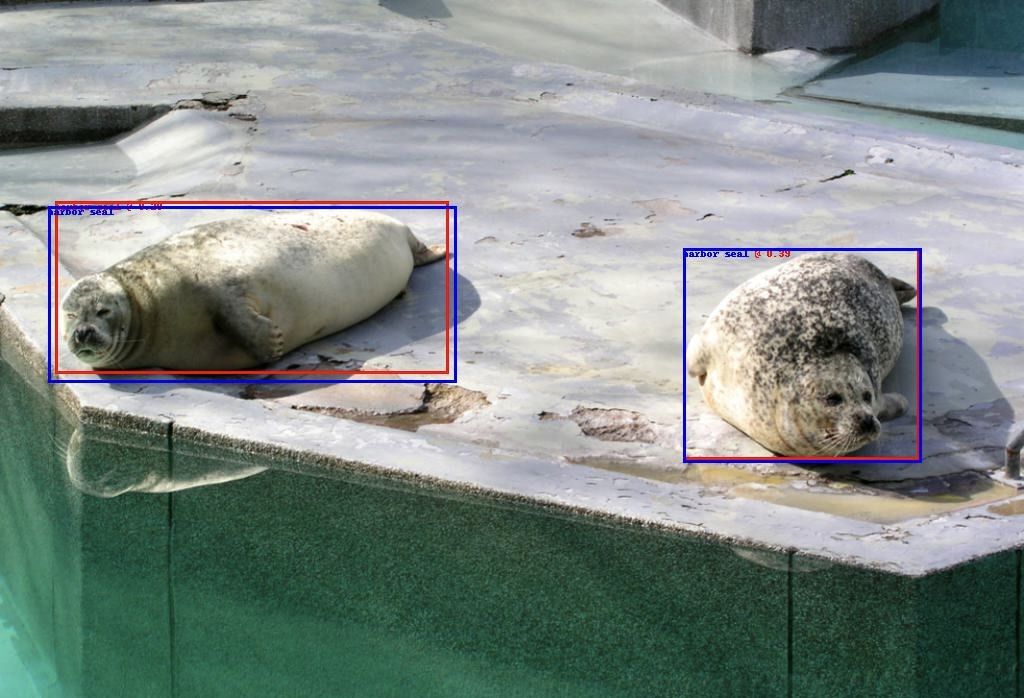}
\end{subfigure}
\begin{subfigure}[b]{0.155\textwidth}
\includegraphics[width=\textwidth]{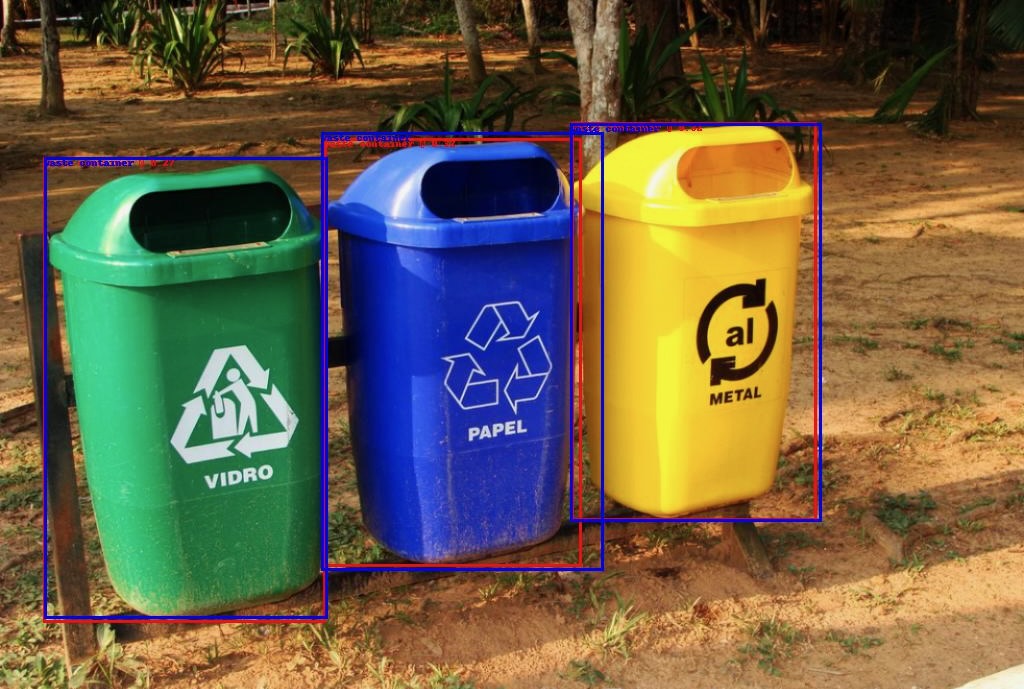}
\end{subfigure}
\begin{subfigure}[b]{0.155\textwidth}
\includegraphics[width=\textwidth]{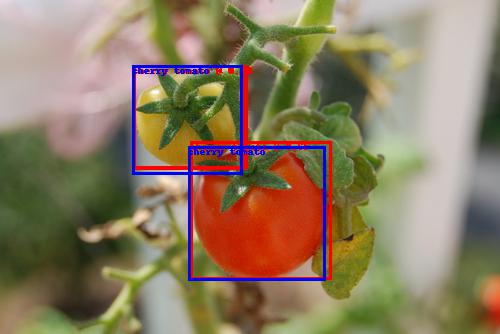}
\end{subfigure}
%\begin{subfigure}[b]{0.12\textwidth}
%\includegraphics[width=\textwidth]{orchid.jpg}
%\end{subfigure}
%\begin{subfigure}[b]{0.12\textwidth}
%\includegraphics[width=\textwidth]{bowling3.jpeg}
%\end{subfigure}
%\begin{subfigure}[b]{0.12\textwidth}
%\includegraphics[width=\textwidth]{tiara3.jpeg}
%\end{subfigure}
\caption{Sample of results of WS-DETR Full on the FSOD-200 held-out test set. Best viewed electronically and zoomed in. Blue represents ground-truth and red represents our WS-DETR prediction. The model has not seen bounding box labels for classes in the test set.}
\label{fig:success}
\end{figure}

\section{Conclusion}
\label{sec:conclusion}
We propose the Weakly Supervised Detection Transformer (WS-DETR), which integrates DETR with an MIL architecture for scalable WSOD finetuning on novel objects. Our hybrid model leverages the strengths of both two-stage detectors and the end-to-end DETR framework. In comparison to existing WSOD approaches, which only operate at small data scales and require multiple rounds of training and refinement, our WS-DETR utilizes a single pretrained model for knowledge transfer to WSOD finetuning in a variety of practical domains. We achieve state-of-the-art performance in novel and fine-grained settings, and our scaling study reveals that class quantity is more important than image quantity for WSOD pretraining. 

\textbf{Potential negative social impact.} Object detection models have malicious potential for surveillance. Ours could have unintended negative impact by lessening the labeling needed to detect fine-grained categories of people; we explicitly discourage these applications. Additionally, the environmental cost of pretraining Transformers on massive datasets is significant, so we will release our checkpoints for others to utilize with minimal added emissions.

%\newpage

{\small
\bibliographystyle{ieee_fullname}
\bibliography{refs}
}

\end{document}